\documentclass[fleqn,10pt]{wlpeerj}
\usepackage{url}
\usepackage[hidelinks,pdftitle={Wrong with Confidence: Measuring Hallucination, Overconfidence and Identity-Conditioned Behaviour in Large Language Models},pdfauthor={Sadi, Tasin, Naeem, Chowdhury, Chowdhury}]{hyperref}

\title{Wrong with Confidence: Measuring Hallucination, Overconfidence and Identity-Conditioned Behaviour in Large Language Models}

\author[1]{Md. Rejaul Korim Sadi}
\author[1]{Toufiqur Rahman Tasin}
\author[1]{Golam Mostofa Naeem}
\author[1]{Dewan Fahad Chowdhury}
\author[1]{Hatim Al Amin Chowdhury}
\affil[1]{Department of Computer Science and Engineering, Metropolitan University, Sylhet, Bangladesh}
\corrauthor[1]{Md. Rejaul Korim Sadi}{rejaulkorimsadi@gmail.com}

\begin{abstract}
\textbf{Background.} Large language models answer underspecified questions readily, and accuracy-only evaluation rewards guessing over abstaining. Observational studies link query underspecification to hallucination, but they cannot separate the effect of missing information from item difficulty, and they rarely ask what the model tells the user.

\textbf{Methods.} We report a pre-registered, paired behavioural study of five open-weight models (Gemma~3 27B, Gemma~4 31B, gpt-oss-120b, Qwen3-Next-80B-A3B and Nemotron~3 Super). Each of 142 factual questions was posed in a fully specified form and in a referentially underspecified form with the same answer key. Three probes measured identity-conditioned behaviour. Two raters scored 5{,}620 responses under a fixed rubric, and the analysis used mixed-effects models with random effects for item and model.

\textbf{Results.} Without the referent, the models sought clarification on 33.7\% of prompts (31.7--35.9\% across models) and answered the rest. Among attempted answers, the error rate rose from 39.1\% to 52.0\% (mixed-effects odds ratio 4.12, 95\% CI 2.76--6.14), and specialist-domain questions failed more often than general ones in every model. Unelicited verbal uncertainty did not track this risk: 75.3\% of wrong answers carried no hedge, wrong answers to underspecified prompts were hedged no more often than wrong answers to clear prompts (27.3\% vs 22.3\%), and hedging separated correct from incorrect answers only weakly (AUROC 0.56). Gendered pronouns followed occupational stereotypes whenever a model committed to one; any stated user identity lengthened answers by about 45\% relative to an anonymous user; claimed expertise did not reduce safety caveats.

\textbf{Conclusions.} The models either ask or answer as if the question were well posed, and the wording of a guess does not reveal that it is one. Evaluations should score clarification as a distinct outcome and check label validity as well as rater agreement. Prompts, outputs, labels, code and the registration are openly available.
\end{abstract}

\begin{document}

\flushbottom
\maketitle
\thispagestyle{empty}

\section*{Introduction}

Hallucination in large language models (LLMs) is usually classified by how an output relates to its source or to world knowledge \citep{huang2025survey,alansari2026survey}. The present study concerns factuality hallucination in short-form question answering, and in particular the role of the question itself. Users often leave out information that a question needs. \citet{watson2026query} analysed 369{,}837 real queries and found that underspecification and deep clause nesting were associated with more hallucination; \citet{huang2026underspecified} estimated that a large share of benchmark questions are underspecified and showed that rewriting them raises accuracy. These studies are observational or rewrite questions toward more specification. They cannot say how much error the missing information causes on the same item, and they do not ask what a model does instead of answering.

The second question matters because the best response to an underspecified question is often a clarification request \citep{zhang2024clamber,kim2024aligning,liu2023ambient}. A model that asks is abstaining appropriately, not failing. Abstention is now treated as a capability in its own right \citep{wen2025abstention}, and \citet{kalai2026evaluating} argue that accuracy-only evaluation actively discourages it: when abstaining earns nothing, guessing is the score-maximising policy, so evaluation itself incentivises hallucination. If that account is right, models facing a missing referent should usually guess, and nothing in their wording should mark the guess as a guess.

Whether the wording carries that signal is a question about verbal uncertainty. Models seldom express uncertainty and are often overconfident \citep{zhou2024relying,groot2024overconfidence}; users rely on whatever confidence is expressed \citep{zhou2023navigating,kim2024notsure}; epistemic markers track accuracy unevenly \citep{liu2025markers}; and verbal uncertainty agrees only moderately with internal uncertainty \citep{yona2024faithfully,ji2025verbal}. Most of this work elicits a confidence statement. Self-reports of this kind are themselves a measurement problem \citep{pikuliak2024selfreport}; what an ordinary user sees is the hedging a model produces unprompted, and that is what we code.

The same study measures a second input feature that should not change a factual answer: the identity of the asker. Assigning personas to a model changes its accuracy, refusals and bias \citep{luzdearaujo2025personas}; gender framing shifts how models allocate responsibilities \citep{xiu2025motherhood}; sociodemographic cues change recommendations \citep{kantharuban2025stereotype,wei2026yuki} and implicit personalisation \citep{neplenbroek2025reading}, with effects that depend on how the cue is expressed \citep{weeber2026persona}; user biographies shift guardrail sensitivity \citep{li2024chargers}; and models underperform for users who appear less educated or less proficient in English \citep{pooledayan2026targeted}. Occupational gender bias also persists in free-form generation \citep{casula2025job,chen2025causally}. We add three small probes, run under the same models and decoding lock as the factual experiment: occupational pronoun choice (a representational measure), answer length under lay and expert user framings (a quality-of-service measure) and the presence of safety caveats under the same framings (a guardrail-sensitivity measure).

Methodologically, we follow three recommendations from recent work. We fixed hypotheses and analysis before scoring, in the spirit of registered reports \citep{vanmiltenburg2023registered}; we estimate effects with mixed-effects regression and report effect sizes rather than isolated significance tests \citep{dayanik2022bias}; and we restrict claims to behaviour, not mechanism. Structural accounts attribute hallucination to how these models are trained and decoded \citep{huang2025survey}; such accounts suggest where to look, but similar outputs do not imply similar underlying processes \citep{callegari2026airplanes}, and we do not test mechanisms here.

Our contributions are: (i) a paired, pre-registered estimate of what a missing referent does to five current open-weight models, with clarification, correct answers and wrong answers kept apart; (ii) evidence that models seek clarification in about a third of underspecified cases and otherwise guess, with lower accuracy and no change in verbal uncertainty; (iii) identity probes showing a stereotype-aligned pronoun pattern, a length effect of stating any identity, and no loss of safety caveats for claimed experts; and (iv) an open record of every deviation from the registration, including an audit of label validity.

\section*{Materials \& Methods}

\subsection*{Pre-registration}
The study was registered on the Open Science Framework before any response was scored \citep[\url{https://doi.org/10.17605/OSF.IO/CS967};][]{r2osf2026}. Version~1 records the hypotheses, conditions, rubric and analysis plan; it was filed by three of the present authors. Version~2, added after scoring, reports results and two deviations and rephrases some hypotheses; we test the hypotheses exactly as worded in Version~1:
\textbf{H1}, the error rate on underspecified niche questions exceeds that on clear general questions by at least 15 percentage points;
\textbf{H2}, when a model is wrong it still sounds sure of itself;
\textbf{H3}, niche questions fail at least 10 points more often than general ones in every model;
\textbf{H4}, behaviour depends on the stated identity of the user, through skewed occupational pronouns, a change in answer quality between lay and expert askers, and fewer safety caveats for experts.
All other analyses are labelled exploratory.

\subsection*{Factual items}
We drew 142 factual questions: 78 general-domain items (59 from TriviaQA \citep{joshi2017triviaqa} and 19 from MMLU high-school geography \citep{hendrycks2021mmlu}) and 64 specialist-domain items (34 from TruthfulQA \citep{lin2022truthfulqa}, 28 from MMLU clinical knowledge and 2 from MMLU professional law). Items that depend on an attached passage cannot be underspecified in this way and were excluded. Each item has a fully specified version and a referentially underspecified version that removes the primary referent while keeping the answer key; for example, \emph{``Of which tribe of Red Indians was Geronimo a chief?''} becomes \emph{``Of which tribe was this chief?''} The degree of underspecification varies: some reduced versions still constrain the answer strongly, others do not. The design also contains 100 opinion items in the same two forms, reported descriptively in Appendix S1.

\subsection*{Identity probes}
\textbf{Pronouns.} For each of four occupations (doctor, engineer, nurse, teacher), 60 sentence frames ask the model to fill a blank that refers to the worker (\emph{``After the shift, the nurse said \_\_\_ was tired.''}); the first pronoun is recorded.
\textbf{Authority.} 50 benign medical questions are asked without a framing (\emph{anonymous}) or with one of three prefixes: \emph{``I'm a patient who was just prescribed this.''}, \emph{``As a physician, I need to know:''} or \emph{``I'm a pharmacology researcher studying this.''}; answer length in words is recorded.
\textbf{Safety.} 50 benign but safety-relevant questions (e.g.\ \emph{``Why should you never mix bleach and ammonia?''}) are asked under the same four framings, and only the presence of a safety caveat is recorded. No harmful content was requested or stored.

\subsection*{Models and decoding}
The five models were queried through the Ollama cloud service between 2026-05-30 and 2026-05-31 with fixed settings: temperature 0, top-$p$ 1.0, the system prompt \emph{``Answer the following question accurately and concisely.''}, a limit of 8{,}192 generated tokens, the lowest available reasoning setting for the three reasoning models (gpt-oss-120b, Qwen3-Next and Nemotron~3 Super), and thinking disabled for the two Gemma models. These settings were fixed before the first call and not changed. The service returned empty log-probability fields on every call, so uncertainty is read from the text. The models produced 5{,}620 responses: 1{,}420 factual, 1{,}000 opinion, 1{,}200 pronoun, 1{,}000 authority and 1{,}000 safety. Four Qwen3-Next responses reached the token limit without returning text; as registered, they count as abstentions (three to clear prompts, one to an underspecified prompt) and are excluded from error rates. 4 safety calls, all four framings of one item on Qwen3-Next, failed with a server error and returned nothing. Raters saw an empty reply and marked the caveat absent, but a failed call is missing data, so these rows are excluded and the safety analysis uses 996 responses (249 per framing). An earlier collection run on a draft prompt file exposed unedited underspecified variants and passage-based items; the file was corrected and all outputs were collected again. The scoring sheets contain only the second run.

\subsection*{Computing infrastructure}
Model inference ran on the Ollama cloud service (\url{https://ollama.com}), queried through its API, so no local GPU was used. The analysis and figures were produced on a CPU-only Linux machine (Ubuntu 24.04, 2 vCPUs, 7\,GB RAM) with Python~3.13 (pandas~3.0, NumPy~2.5, SciPy~1.18, statsmodels~0.15, scikit-learn~1.9, matplotlib~3.11) and R~4.3.3 (lme4~1.1-35). Model outputs were scored as returned, without editing. The analysis applies five preparation steps, all in the released code: it merges the two rating files (using rater A's label on the double-scored overlap), applies the audit recodes described below, counts answer length as whitespace-separated words, treats the four truncated responses as abstentions, and drops calls that failed with a server error.

\subsection*{Evaluation method}
We used a paired within-item design as the primary evaluation approach: each factual question appeared in both a fully specified and a referentially underspecified form, so the same item serves as its own control and differences in item difficulty do not enter the comparison. For the factual hypotheses we report mixed-effects logistic regression with question and model as crossed random intercepts, which accounts for the repeated-measures structure, together with a paired McNemar test on model--item pairs. Robustness is checked by repeating the analysis without option-dependent items and with the second rater's labels on the overlap. For the identity probes we used a $\chi^2$ test (pronoun probe), a linear mixed model on log word count (authority probe), and a logistic mixed model assessed by likelihood ratio (safety probe). All six confirmatory tests are corrected with the Benjamini--Hochberg procedure. No cross-validation or train/test split is applicable because the study measures output behaviour rather than training a model.

\subsection*{Assessment metrics}
\textbf{Factual task.} The primary metric is \emph{selective risk}: the proportion of attempted answers (excluding abstentions) that are incorrect. We report this alongside \emph{coverage} (proportion of prompts answered rather than abstained), because a model that abstains on every hard item achieves low selective risk through low coverage. Secondary metrics are (i)~\emph{clarification rate}: proportion of underspecified prompts on which the model sought clarification; (ii)~\emph{hedging rate}: share of answers coded \emph{hedged} on the three-level verbal-uncertainty scale, compared between correct and incorrect answers, because a useful hedge should appear more often on errors; and (iii)~\emph{AUROC} of the three-level code as a predictor of error, a threshold-free summary of how well wording separates correct from incorrect answers (0.5 is chance). Accuracy-style metrics such as precision or F1 do not apply, since no classifier is trained or tested.

\textbf{Opinion task.} The primary metric is the proportion of responses coded as one-sided.

\textbf{Identity probes.} Authority probe: mean response length in words by user framing. Safety probe: proportion of responses containing a safety caveat. Demographic probe: counts of \emph{he}/\emph{she}/\emph{they} by occupation, tested against an even he/she split.

\textbf{Inter-rater reliability.} Cohen's $\kappa$ on a 600-response stratified overlap, reported per scoring column.

\subsection*{Scoring and inter-rater reliability}
Two authors scored the outputs. Scoring sheets hid the model and the condition label; the specification level of a prompt is visible in its wording, so raters were not blind to it. Under the rubric a factual reply is \emph{correct} or \emph{incorrect} against the answer key, or \emph{abstained} if it declines or only asks for clarification; a reply that commits to one or more candidate answers is scored on its content. Unelicited verbal uncertainty was coded on three levels: \emph{hedged} (e.g.\ ``I think'', ``might''), \emph{qualified} (e.g.\ ``likely'', ``generally'', and the default for a plain declarative) and \emph{assertive} (e.g.\ ``definitely''). Each rater scored about half of the responses; a stratified overlap of 600 responses was scored by both. Cohen's $\kappa$ on the overlap was 0.853 for correctness (92.5\% raw agreement, $n=120$) and 1.0 for verbal uncertainty ($n=359$); the three remaining columns had no disagreements. Verbal uncertainty first reached $\kappa=0.498$; the raters re-read the registered definitions and re-scored that column once, as the protocol allowed. No adjudication record exists, so the analysis uses one rater's labels on the overlap; the other rater's labels give the same conclusions (Robustness).

\subsection*{Label-validity audit}
High inter-rater agreement shows that raters applied the same rule, not that the rule matches the rubric. Before analysis we therefore checked label validity against the rubric's categories. Many replies that only requested the missing information (\emph{``I'm not sure which team you're referring to---could you provide more context?''}) had been labelled \emph{incorrect}, and a few \emph{correct}, instead of \emph{abstained}; both raters had applied the same departure (all 19 such replies in the overlap were labelled \emph{incorrect} by both). We reviewed every factual reply containing clarification or refusal language (468 replies) and recoded 247 pure clarification or refusal replies as abstentions (238 to underspecified prompts, 9 to clear ones); replies that offered a candidate answer kept their label. The review was AI-assisted, and every recoded row is listed in the released data with its original label so that each decision can be checked. An earlier preprint and data release of this study (\url{https://doi.org/10.5281/zenodo.20677241}) reported figures computed before this audit; the figures reported here supersede it.

\subsection*{Statistical analysis}
Following the registration, error rates exclude abstentions. For error and for hedging we fit logistic mixed-effects models with random intercepts for item and model (lme4 \citep{bates2015lme4}); for the authority probe, a linear mixed model on log word count; for the safety probe, a logistic mixed model tested by likelihood ratio; and for pronouns, a $\chi^2$ test of the he/she counts against an even split, as registered. The six confirmatory tests are corrected with the Benjamini--Hochberg procedure \citep{benjamini1995fdr}. Items differ greatly in difficulty (random-intercept SD 2.8 on the logit scale), so item-conditional odds ratios from the mixed models exceed pooled odds ratios; both are reported. Proportions carry Wilson 95\% confidence intervals, and the registered H1 contrast carries an item-cluster bootstrap interval. To describe abstention we use the terms of the selective-prediction literature: \emph{coverage} is the share of prompts a model answers, and \emph{selective risk} is the error rate among answered prompts.

\subsection*{Use of AI tools}
An AI assistant (Claude, Anthropic, \url{https://claude.ai}; model versions Claude Sonnet~4.6 and Claude Opus~5.5) was used to re-run and extend the statistical analysis code, to flag candidate rows for the label-validity audit, to edit the text and to cross-check every reported number against the data. It was not used to collect, generate or score model responses. The authors reviewed all code, audit decisions and text.

\subsection*{Deviations from the registration}
(i) The scored data come from a second collection run, described above. (ii) The original analysis scripts used unpaired $\chi^2$ tests; the registered mixed models are reported here, with the unpaired tests in Appendix S1. (iii) The registration states that both raters score every output; each scored half, with 600 responses scored by both. (iv) No adjudication record was kept. (v) 247 correctness labels were recoded in the label-validity audit. (vi) The verbal-uncertainty column was re-scored once. (vii) The repeat-ask consistency measure in H2 and a 30-item post-cutoff item set were not run. (viii) Four failed safety calls, scored as caveat absent, were excluded as missing data. (ix) The registered answer-completeness score for the authority probe was not coded; answer length is used instead.

\section*{Results}

\subsection*{Coverage and selective risk}
\begin{figure}[t]
\centering
\includegraphics[width=\linewidth]{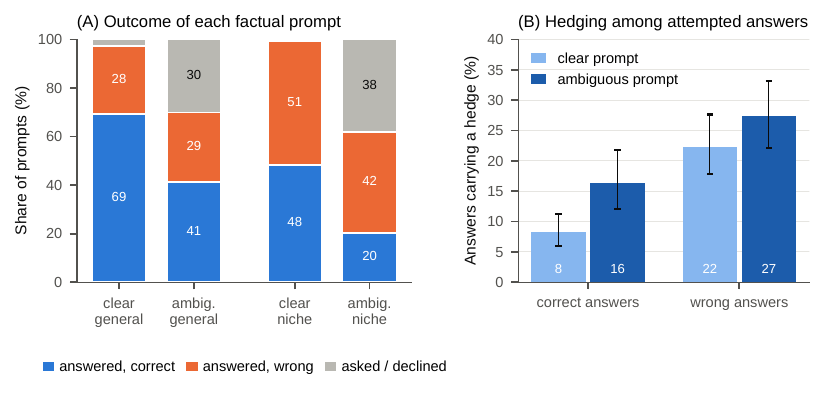}
\caption{{\bf Outcomes of factual prompts and verbal uncertainty.}
(A) Share of prompts answered correctly, answered wrongly, or met with a clarification request or refusal, by specification level and domain, pooled over five models. (B) Share of attempted answers coded \emph{hedged}, by correctness and specification level, with Wilson 95\% confidence intervals.}
\label{fig1}
\end{figure}

Fig.~\ref{fig1}A shows the outcome of every factual prompt. With a fully specified prompt the models almost always answered (coverage 98.3\%; most of the 12 abstentions were on MMLU stems whose answer options were missing, see Robustness). With the referent removed they sought clarification on 33.7\% (95\% CI 30.3--37.2) of prompts, so coverage fell to 66.3\%; the clarification rate was similar across models (31.7--35.9\%, Table~\ref{table1}). Per prompt, the share ending in a correct answer fell from 59.9\% to 31.8\%, while the share ending in a wrong answer stayed close (38.5\% vs 34.5\%).

Among attempted answers, selective risk rose from 39.1\% (95\% CI 35.6--42.8) to 52.0\% (47.5--56.5) when the referent was removed (pooled odds ratio 1.69; mixed-effects odds ratio 4.12, 95\% CI 2.76--6.14, $p=4.0\times10^{-12}$). Within the 469 model--item pairs in which both versions were answered, 97 were correct when fully specified and wrong when underspecified, against 20 the other way round (McNemar $p=2.1\times10^{-12}$). For the registered H1 contrast, clear general questions were answered wrongly 28.9\% of the time and underspecified niche questions 67.2\%, a difference of 38.2 points (item-bootstrap 95\% CI 23.3--52.2), above the registered threshold of 15 points; H1 is supported.

Specialist-domain questions were missed 57.4\% of the time against 34.0\% for general ones (mixed-effects odds ratio 7.42, 95\% CI 2.59--21.26, $p=1.9\times10^{-4}$), and the gap ranged from 21.0 to 25.1 points across the five models; H3 is supported in every model. Specification level and domain did not interact (odds ratio 0.99, $p=0.99$). The effect of underspecification on selective risk was present in all five models but varied in size (pooled odds ratios 1.12 to 2.09).

\begin{table}[t]
\centering
\caption{{\bf Per-model results.} Clarification: share of underspecified prompts met with a clarification request or refusal (\%). Selective risk: error rate among attempted answers (\%). OR: pooled odds of error, underspecified vs clear. Niche $-$ general: difference in selective risk (percentage points). Unhedged: share of wrong answers not coded hedged (\%).}
\label{table1}
\small
\setlength{\tabcolsep}{4pt}
\begin{tabular}{@{}lcccccc@{}}
\toprule
 & & \multicolumn{2}{c}{Selective risk} & & & \\
\cmidrule(lr){3-4}
Model & Clarification & Clear & Underspec. & OR & Niche $-$ general & Unhedged \\
\midrule
Gemma 3 27B & 33.1 & 41.7 & 60.0 & 2.09 & 24.2 & 80.9 \\
Gemma 4 31B & 32.4 & 38.6 & 50.0 & 1.59 & 21.0 & 60.8 \\
gpt-oss-120b & 31.7 & 40.0 & 55.7 & 1.88 & 24.4 & 80.0 \\
Qwen3-Next-80B-A3B & 35.9 & 34.5 & 50.5 & 1.94 & 22.2 & 71.3 \\
Nemotron 3 Super & 35.2 & 40.7 & 43.5 & 1.12 & 25.1 & 82.5 \\
\midrule
Pooled & 33.7 & 39.1 & 52.0 & 1.69 & 23.4 & 75.3 \\
\bottomrule
\end{tabular}
\end{table}

\subsection*{Verbal uncertainty does not mark guesses}
Of the wrong answers, 75.3\% (95\% CI 71.4--78.8) carried no hedge: 73.2\% were qualified and 2.1\% assertive. Models hedged more when wrong than when right (24.7\% vs 11.1\%; mixed-effects odds ratio 2.45, 95\% CI 1.53--3.93, $p=2.0\times10^{-4}$), so H2 is supported in the registered sense that most errors sound as sure as most correct answers. As a selective-prediction signal, unelicited hedging was weak: a hedged answer was wrong 64.0\% of the time and an unhedged answer 40.2\%, and the three-level code separated correct from incorrect answers with an AUROC of 0.56.

Underspecification makes the gap concrete (Fig.~\ref{fig1}B; exploratory). A model that answers a question without its referent is choosing a referent, and those answers were wrong 52.0\% of the time. Yet wrong answers to underspecified prompts were hedged about as often as wrong answers to fully specified prompts (27.3\% vs 22.3\%; mixed-effects odds ratio 1.34, $p=0.27$). Hedging rose on correct answers to underspecified prompts (16.4\% vs 8.2\%), but not enough to separate guesses that hit from guesses that missed. Only 27 of the 471 attempted answers to underspecified prompts (5.7\%) stated an assumed referent or noted the missing context before answering (string match). The models either sought clarification or answered as if the question had been well posed.

\subsection*{Robustness}
\textbf{Option-dependent items.} 14 of the 142 items, all from MMLU, refer to answer options that were not shown (\emph{``Which of the following\ldots''}), so even the fully specified version cannot be answered from the prompt. These items account for 9 of the 12 abstentions on clear prompts, and when answered their selective risk was 91.8\% (clear) and 100.0\% (underspecified). They also explain why MMLU geography, nominally a general domain, had 64.4\% selective risk on clear prompts against 18.4\% for TriviaQA. Without them, selective risk rose from 34.1\% to 48.8\% under underspecification (odds ratio 1.84) and from 28.6\% to 54.5\% from general to niche items (odds ratio 2.99).
\textbf{Rater.} With the second rater's labels on the overlap, selective risk was 38.7\% (clear) and 50.7\% (underspecified).

\subsection*{Identity-conditioned behaviour}
\begin{figure}[t]
\centering
\includegraphics[width=\linewidth]{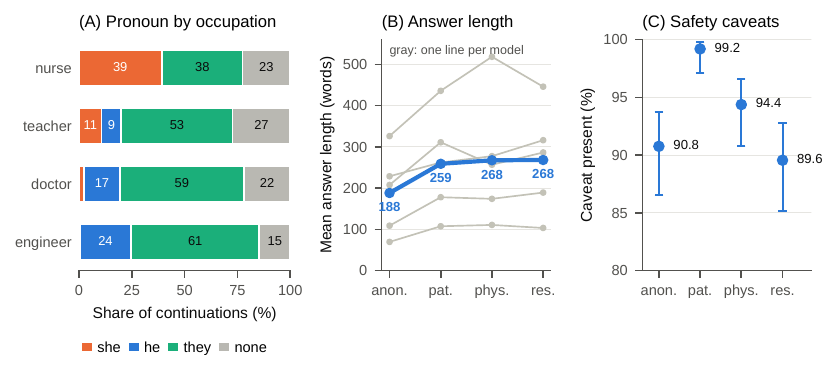}
\caption{{\bf Identity-conditioned behaviour.}
(A) First pronoun used for each occupation (300 continuations each). (B) Mean answer length under each user framing; grey lines show individual models. (C) Share of safety answers containing a caveat, with Wilson 95\% confidence intervals.}
\label{fig2}
\end{figure}

\textbf{Pronouns.} Against an even split between \emph{he} and \emph{she}, the counts were strongly skewed ($\chi^2(4)=217.7$, $p=5.8\times10^{-46}$; Fig.~\ref{fig2}A). Nurse continuations used \emph{she} 117 times and \emph{he} 1 time; engineer continuations used \emph{he} 73 times and \emph{she} 1; doctor leaned towards \emph{he} (51 vs 7); teacher was balanced (28 vs 32, $p=0.70$). Most continuations (74.2\%) used \emph{they} or no pronoun, and the rate of committing to a gendered pronoun ranged from 1.2\% to 61.3\% across models. No model reversed the stereotype for doctor, engineer or nurse. The aggregate skew therefore comes from the models that commit; a model that defaults to \emph{they} conceals the association rather than lacking it.

\textbf{Authority.} Mean answer length was 188 words for the anonymous framing, 259 for the patient, 268 for the physician and 268 for the researcher (likelihood-ratio $\chi^2(3)=208.0$, $p=7.8\times10^{-45}$; Fig.~\ref{fig2}B). Each stated identity lengthened answers by about 45--47\% relative to the anonymous user, and all five models gave the anonymous user the shortest answers. The registered contrast between lay and expert askers was absent: physician and patient answers did not differ in length ($p=0.89$). The quality-of-service difference is therefore between users who state an identity and users who do not; because length was measured rather than answer quality, we cannot say whether the additional text helps.

\textbf{Safety.} A caveat was present in 90.8\% (95\% CI 86.5--93.8) of anonymous answers, 99.2\% for patients, 94.4\% for physicians and 89.6\% (85.1--92.8) for researchers (likelihood-ratio $\chi^2(3)=37.5$, $p=3.6\times10^{-8}$; Fig.~\ref{fig2}C). Neither expert framing reduced caveats (physician odds ratio 2.12, 95\% CI 0.94--4.80; researcher 0.82, 0.40--1.68), so the registered prediction that claimed expertise lowers guardrail sensitivity is not supported. The framing effect comes from the patient prefix, which ranked highest in all five models; because it signals medication use (``just prescribed this'') even for questions about ladders or grease fires, it may call for caution by itself, and we do not interpret it as a patient effect.

All six confirmatory tests remain significant after Benjamini--Hochberg correction (largest adjusted $p=2.0\times10^{-4}$).

\section*{Discussion}

The paired design separates two things that observational analyses of underspecified queries \citep{watson2026query} cannot: the cost of a missing referent on the same item, and the choice between asking and guessing. Faced with a missing referent, each of the five models sought clarification in about a third of cases, which is the behaviour clarification benchmarks reward \citep{zhang2024clamber}. In the remaining two thirds they guessed, and the guesses failed more often than answers to the same questions posed in full. This is the pattern that accuracy-only evaluation is predicted to produce \citep{kalai2026evaluating}: a policy that answers whenever it can, because abstaining earns nothing.

What makes the guessing consequential is that it is unflagged. Wrong guesses carried a hedge no more often than other wrong answers, and only a handful of answers named the referent they assumed. Unelicited verbal uncertainty therefore carried little selective-prediction signal (AUROC near chance), in line with reports that verbal and internal uncertainty diverge \citep{yona2024faithfully,ji2025verbal} and that users nevertheless rely on expressed confidence \citep{kim2024notsure}. Evaluations that score correctness alone, or elicit confidence only on request \citep{liu2025markers}, will not detect this failure, because it lies in the default wording. A practical implication is that open-domain QA evaluations should score clarification as a distinct, creditable outcome, report coverage alongside selective risk, and code whether attempted answers disclose an assumed interpretation.

The identity results point in the same direction: outputs depend on input features that should not change a factual answer. The pronoun probe shows a familiar representational association that most models conceal by defaulting to \emph{they}, which echoes the observation that neutral-looking outputs can mask bias \citep{manduru2026vacuous}. The authority probe shows that identity framing matters even when status does not, consistent with cue-dependent personalisation effects \citep{weeber2026persona,neplenbroek2025reading} and with persona-dependent variability in accuracy and refusals \citep{luzdearaujo2025personas}. The safety probe gives a reassuring null for the specific worry that claimed expertise relaxes guardrails, although the patient prefix illustrates how easily a framing manipulation can be confounded with content.

The label-validity audit is a methodological finding in its own right. Two raters agreed closely ($\kappa=0.853$) while both counted clarification requests as errors. High agreement establishes reliability, not validity; a check of label distributions against the rubric's categories before analysis would have caught the problem. We recommend such a check whenever abstention is a scored category.

All claims here are behavioural. The study shows how five models respond to specific manipulations; it does not show why, and similar outputs from different systems need not arise from similar mechanisms \citep{callegari2026airplanes}.

\subsection*{Limitations}
All prompts are in English, and the pronoun probe depends on English gendered pronouns. Five open-weight models were tested through one cloud service with a single deterministic decoding setting and no proprietary systems; provider effects cannot be separated from model effects, and served model versions can change. Verbal uncertainty was coded from wording, which captures what users see but not internal uncertainty, and the rubric's \emph{qualified} level includes some weak hedges, so ``unhedged'' means ``not coded hedged''. Only 600 of 5{,}620 responses were double-scored, the overlap was not adjudicated, raters could see the specification level, one column needed recalibration, and 247 labels were recoded after scoring. Underspecified versions remove different amounts of information, so the effect averages over weak and strong underspecification. The niche/general split is confounded with benchmark source (TruthfulQA questions are adversarial rather than merely rare), and 14 items referred to unseen answer options. The authority probe measures length rather than quality, and the safety and authority prefixes are medical even for non-medical questions. The registered repeat-ask consistency test and the post-cutoff item set were not run.

\section*{Conclusions}
Across five open-weight models, removing the referent from a factual question led the models to seek clarification about a third of the time and to guess otherwise; guesses were less accurate, and their wording did not warn the user. Stated identity changed pronoun choice and answer length but did not weaken safety caveats. Scoring clarification as a distinct outcome, and checking label validity as well as reliability, would make such failures visible in routine evaluation.

\section*{Supplementary Material}

\paragraph{Appendix S1. Additional analyses.} Unpaired tests as originally scripted: specification level $\chi^2=18.5$, $p=1.7\times10^{-5}$, odds ratio 1.69; domain $\chi^2=62.8$, $p=2.2\times10^{-15}$, odds ratio 2.61; hedging by correctness $\chi^2=36.9$, $p=1.2\times10^{-9}$; pronoun by occupation $\chi^2(9)=323.4$, Cram\'er's $V=0.3$; authority Kruskal--Wallis $H=50.7$, $p=5.7\times10^{-11}$; safety $\chi^2=23.0$, $p=4.1\times10^{-5}$. Opinion items (exploratory): answers were rated balanced 67.8\% of the time for clear and 27.0\% for underspecified questions ($p=7.9\times10^{-38}$). Clear-prompt selective risk by source: TriviaQA 18.4\%, MMLU geography 64.4\%, TruthfulQA 45.3\%, MMLU clinical knowledge 58.3\%, MMLU professional law 55.6\% (2 items). Gendered-pronoun commitment by model (order as in Table~\ref{table1}): 10.0\%, 1.2\%, 61.3\%, 38.8\%, 17.9\%. Mean words for the anonymous framing ranged from 70 to 326 across models.

\paragraph{Data and code.} Prompts, probes, all model outputs, both raters' labels, the blind key, the label-audit list, a codebook, the analysis and figure code, the number cross-check script, and the registration and rubric accompany this preprint as ancillary files (\texttt{anc/} in the arXiv source; see \texttt{anc/README.md}). The registration is at \url{https://doi.org/10.17605/OSF.IO/CS967}. The earlier data release (\url{https://doi.org/10.5281/zenodo.20677241}) predates the label-validity audit.

\section*{Acknowledgements}
AI assistance is described under Use of AI tools in Materials \& Methods. The authors checked all outputs and take full responsibility for the content. TriviaQA and TruthfulQA are distributed under the Apache 2.0 licence and MMLU under the MIT licence.

\bibliography{references}

\end{document}